\documentclass[letterpaper]{article} 
\usepackage{aaai24}  
\usepackage{times}  
\usepackage{helvet}  
\usepackage{courier}  
\usepackage[hyphens]{url}  
\usepackage{graphicx} 
\usepackage{natbib}  
\usepackage{caption} 
\usepackage{algorithm}
\usepackage{algorithmic}
\usepackage{amsmath}
\usepackage{amsfonts}
\usepackage{makecell}
\usepackage{xcolor}
\usepackage{lipsum}

\DeclareMathOperator*{\argmin}{arg\,min}
\usepackage{newfloat}
\usepackage{listings}
\DeclareCaptionStyle{ruled}{labelfont=normalfont,labelsep=colon,strut=off} 
\floatstyle{ruled}
\newfloat{listing}{tb}{lst}{}
\floatname{listing}{Listing}
\title{Compositional Inversion for Stable Diffusion Models}
\author{
    Xulu Zhang\textsuperscript{\rm 1,\rm 2},
    Xiao-Yong Wei\textsuperscript{\rm 3,\rm 1}\thanks{Corresponding Author (x1wei@polyu.edu.hk).},
    Jinlin Wu\textsuperscript{\rm 2,\rm 5},
    Tianyi Zhang\textsuperscript{\rm 1},\\
    Zhaoxiang Zhang\textsuperscript{\rm 2,\rm 4,\rm 5},
    Zhen Lei\textsuperscript{\rm 2,\rm 4,\rm 5},
    Qing Li\textsuperscript{\rm 1}
}
\affiliations{
    \textsuperscript{\rm 1}Department of Computing, the Hong Kong Polytechnic University, Hong Kong\\
    \textsuperscript{\rm 2}Center for Artificial Intelligence and Robotics, HKISI, CAS, Hong Kong\\
    \textsuperscript{\rm 3}College of Computer Science, Sichuan University, Chengdu, China\\
    \textsuperscript{\rm 4}School of Artificial Intelligence, UCAS, Beijing, China\\
    \textsuperscript{\rm 5}State Key Laboratory of Multimodal Artificial Intelligence Systems, CASIA, Beijing, China

}

\usepackage{bibentry}

\begin{document}

\maketitle


\begin{abstract}
Inversion methods, such as Textual Inversion, generate personalized images by incorporating concepts of interest provided by user images. 
However, existing methods often suffer from overfitting issues, where the dominant presence of inverted concepts leads to the absence of other desired concepts. 
It stems from the fact that during inversion, the irrelevant semantics in the user images are also encoded, forcing the inverted concepts to occupy locations far from the core distribution in the embedding space.
To address this issue, we propose a method that guides the inversion process towards the core distribution for compositional embeddings. 
Additionally, we introduce a spatial regularization approach to balance the attention on the concepts being composed. 
Our method is designed as a post-training approach and can be seamlessly integrated with other inversion methods.
Experimental results demonstrate the effectiveness of our proposed approach in mitigating the overfitting problem and generating more diverse and balanced compositions of concepts in the synthesized images.
The source code is available at https://github.com/zhangxulu1996/Compositional-Inversion.
\end{abstract}

\section{Introduction}
\label{sec:intro}
Recently, image synthesis has witnessed remarkable performance from text-to-image diffusion models such as DALL\textbullet E \cite{ramesh2021zero}, Stable Diffusion \cite{rombach2022high}, Imagen \cite{saharia2022photorealistic}.
These models typically consist of two modules: semantic embedding and diffusion.
Given a simple text prompt like ``\textit{a cat chasing butterflies}'', the semantic embedding module represents the semantics as embeddings, while the diffusion module transforms the embeddings into images that incorporate the desired concepts (e.g., \textit{cat}, \textit{butterflies}).
However, these models produce concepts in a general sense, resulting in randomly assigned appearances for the \textit{cat}.
This limitation becomes apparent when users seek specific concepts, such as their own cat.
It raises challenges to these models in the era of pursuing personalized customization.

Textual Inversion (TI) \cite{gal2022image} remains a core technology to address this limitation.
The underlying hypothesis is that an optimal point exists within the embedding space that can represent the semantics of a specific concept, even if it is difficult to describe in words.
TI evaluates the distance of current embedding to the optimal point through back-propagated gradients based on the reconstruction loss from a few sample images provided by the user.
Instead of updating the model weights like in regular training, TI updates the values of the current embedding towards the optimal while keeping the weights fixed.
The post-training feature enables personalization for a significantly wider range of users and researchers, as it demands fewer computational resources compared to the extensive requirement for pretraining or fine-tuning diffusion models.
We hereafter use a star to denote an inverted specific concept (e.g., \textit{cat*}), commonly referred to as a pseudo word in literature.

\begin{figure*}[t]
\centering
\includegraphics[width=0.85\textwidth]{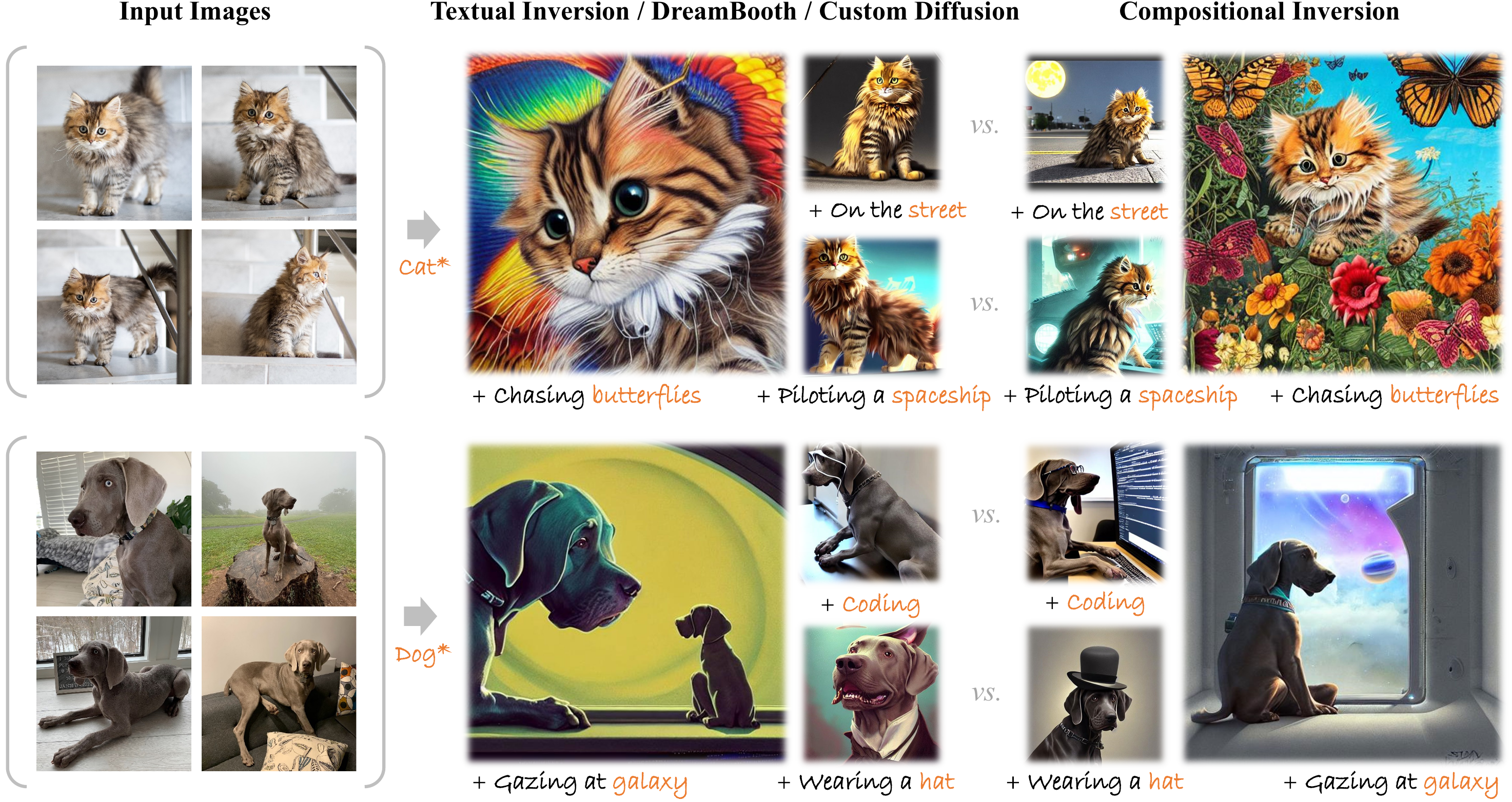} 
\caption{Image synthesis using traditional inversion methods and the proposed compositional inversion: concepts of ${butter}\hspace{-0.075cm}{flies}$, $street$, and $spaceship$ are absent when composed with concepts inverted with traditional methods.}
\label{fig:super_imgs}
\end{figure*}

\begin{figure}[t]
\centering
\includegraphics[width=0.85\columnwidth]{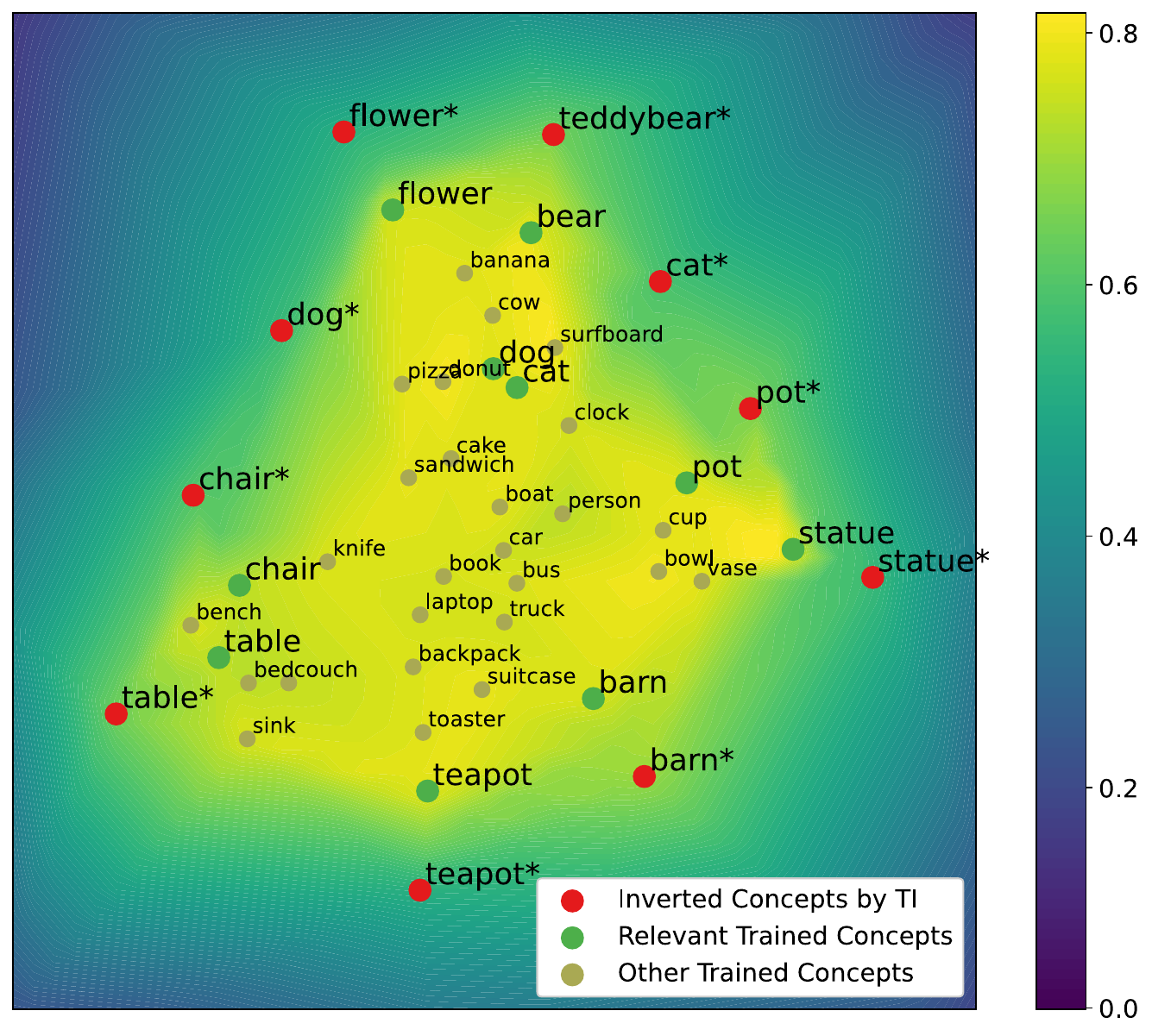} 
\caption{Visualization of compositionality in the embedding space with the evident core distribution and the OOD.}
\label{fig:TSNE}
\end{figure}

\begin{figure}[t]
\centering
\includegraphics[width=0.96\columnwidth]{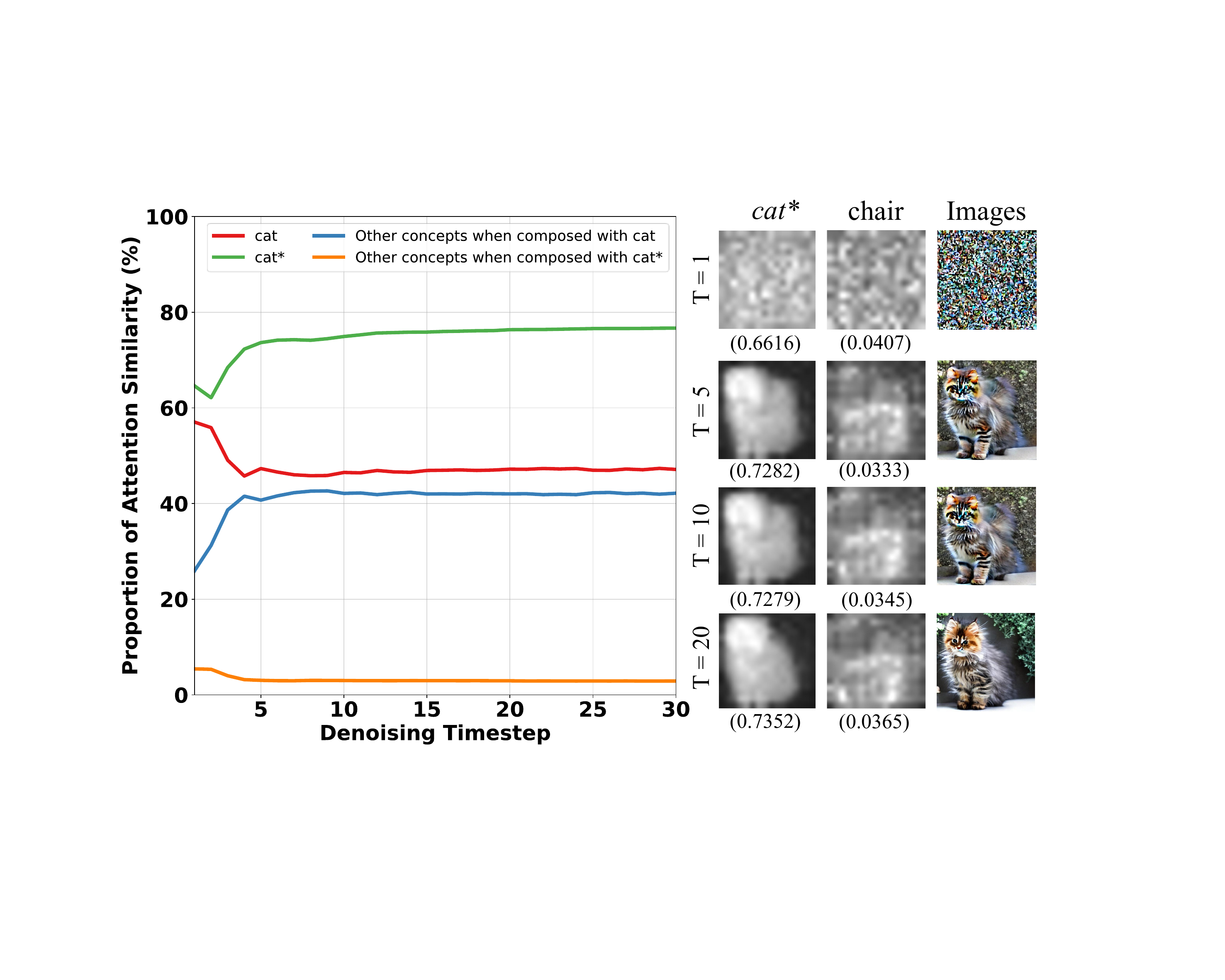} 
\caption{Development of the relative attention similarity and attention maps of various types of concepts.}
\label{fig:attn_sim}
\end{figure}

Despite the presence of promising outcomes, the composition of inverted concepts with other concepts proves to be challenging.
As shown in Fig.~\ref{fig:super_imgs}, the results maintain fidelity to the user samples for \textit{cat*}, but the concept \textit{butterflies} is absent.
This occurs because the method primarily emphasizes the reconstruction loss while disregarding the compositional aspect of the target concept in relation to others.
Similar findings are reported in \cite{tewel2023key} which suggests the dominance of the inverted concepts in the generation process encroaches upon the spotlight of other concepts.
However, this is simply attributed to an over-fitting problem, with underlying rationale remaining unexplored.

This paper represents an initial endeavor to delve into the underlying reasons and offer straightforward solutions from an internal perspective.
Specifically, we have discovered that Textual Inversion leads to the inverted concepts being out of distribution (OOD).
Modern models are always trained on large-scale dataset like LAION \cite{schuhmann2022laion} containing text-image pairs on the billions scale.
Most existing concepts have thus been trained to be compositional to others due to their frequent occurrence in the dataset.
It forms a core distribution where the pretrained concepts are easily combinable.
We have evaluated the compositionality of each concept by combining with others in prompts and testing the probability of their presence in resulting images using object detection.
In Fig.~\ref{fig:TSNE}, the core distribution becomes evident through the visualization of the probabilities based on their coordinates in the embedding space.
This visual representation clearly showcases the OOD issue of the inverted concepts.
The OOD results from the calculation of reconstruction loss that is spanned the entire image rather than the target concept region.
It makes other concepts in the background being ``inverted'', leading to the degradation in the purity of the semantics within the inverted concept.
In Fig.~\ref{fig:TSNE}, due to the distraction of background semantics, the inverted concept \textit{dog*} converges to an OOD area instead of the theoretically more appropriate neighborhood around the concept \textit{dog}.
This observation is further supported by the fact the average entropy of the inverted embeddings has been increased by 3\% from that of the pretrained concepts.

The larger entropy consequently causes the dominance of the inverted concept over others in the diffusion module.
The diffusion module utilizes Transformer blocks to transfer text semantics into visual content, where embeddings are employed to construct the \textbf{K}, and the \textbf{Q} is typically initialized with random noise.
Therefore, the presence of a concept heavily relies on the cross-attention of its embedding to the random noise.
As the larger entropy of the inverted concept implies a broader span of dimensions to store semantics, it may have a higher probability of obtaining greater attention similarity compared to other concepts.
In Fig.~\ref{fig:attn_sim}, we present statistics of the development of the attention similarity between \textbf{K} and \textbf{Q} over iterations.
The divergence between the inverted concept and others is much more pronounced than that between a pretrained concept and others. 
The cross-attention mechanism iteratively integrates the attention map of the inverted concept into other concepts, ultimately resulting in the absence of other concepts or their replacement with the inverted concept (e.g., Fig.~\ref{fig:attn_sim}).

\begin{figure*}[t]
\centering
\includegraphics[width=0.8\textwidth]{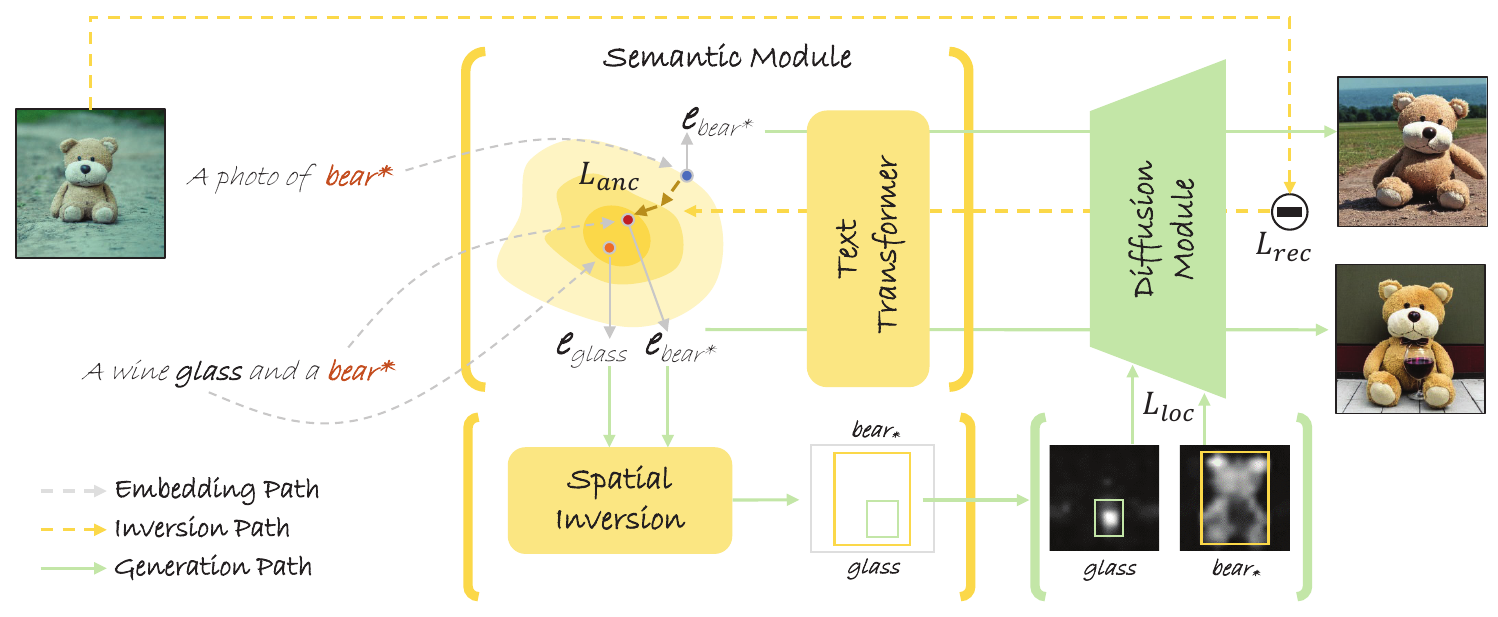} 
\caption{The framework of the proposed method consisting of semantic and spatial inversion components.}
\label{fig:framework}
\end{figure*}

Based on the aforementioned analysis, we propose a compositional inversion approach comprising two components:  a \textbf{semantic inversion} component which guides the embedding search towards the core distribution, and a \textbf{spatial inversion} component which regularizes the attention maps to avoid the dominance of the inverted concepts.
The framework of the proposed method is shown in Fig.~\ref{fig:framework}.


\section{Related Work}
%
Text-to-image synthesis has earned significant attention for its potential applications in content creation, virtual reality, and computer graphics. 
The objective is to bridge the semantic gap and enable machines to understand and generate images that align with text prompts. 
For several years, generative adversarial networks (GANs) \cite{goodfellow2014generative,karras2019style} have been the dominant approach \cite{zhu2019dm,tao2022df}. 
With recent improvements in DDPM \cite{ho2020denoising} and DDIM \cite{song2020denoising}, text-conditioned diffusion models have made remarkable progress. 
Building upon the latent images, the Latent Diffusion Model (LDM) \cite{rombach2022high} was introduced and further extended to Stable Diffusion \cite{rombach2022high}, which is regarded as one of the most promising models for text-to-image synthesis.
Another notable framework, Imagen \cite{saharia2022photorealistic} takes a different approach by diffusing pixels directly using a pyramid structure, without relying on latent images.
DALL\textbullet E2 \cite{ramesh2022hierarchical} uses a prior network that takes text embedding as input to produce an image embedding as the input of the diffusion model.
%
%

\subsection{Inversion for Customization and Personalization}
As aforementioned, it is a demanding feature for the models to generate images containing specific concepts of interest (CoI) implied by user samples.
This requires models' capacity to ``invert'' the samples into concept embeddings, which can be used in future prompts for customized generations.
Textual Inversion \cite{gal2022image} is one of the initial methods that directly searches for the optimal solution in the embedding space to address this issue.
However, the remaining methods, although employing similar approaches of searching for inverted embeddings, rely on either retraining or fine-tuning for this purpose.
For instance, DreamBooth \cite{ruiz2023dreambooth} retrains the entire Imagen for constructing embeddings for CoI, while Custom Diffusion \cite{kumari2023multi}, Perfusion \cite{tewel2023key}, SVDiff \cite{han2023svdiff}, and Cones \cite{liu2023cones} only fine-tune partial parameters of the Stable Diffusion model.
To mitigate language drift and overfitting problems, a large number of images from the same CoI class are typically utilized as regularization during the training/fine-tuning process.

\subsection{Compositionality of Inverted Concepts}
Current methods in the field of compositionality primarily focus on combining inverted concepts with each other rather than with a broader range of pretrained concepts. 
This approach, known as multi-concept composition, is related to but distinct from the scope of this paper. 
Existing methods include Custom Diffusion, SVDiff, Cones, and Perfusion.
Custom Diffusion archives this by merging the outputs of multiple models that have been fine-tuned to invert various CoI. It can be considered as a model-level composition approach.
SVDiff manually combines objects selected from different CoI concepts as training images, enabling the model to learn to compose them.
Cones evaluates the neurons' contributions to the fidelity of inverted concepts and deactivates those with minor contributions during composition.
%
%
Perfusion fuses the \textbf{V} components of inverted concepts to balance their contribution to generation.
These methods all rely on training/fine-tuning, which requires effort and expertise to gather the regularization images.
In contrast, the compositional inversion proposed in this paper is a post-training approach that can be applied to any trained or fine-tuned models and thus is compatible to all the aforementioned methods.
Furthermore, the proposed method can be employed for the composition of both pretrained and inverted concepts, making this paper a study of compositionality in a more general sense.

\subsection{Spatial Guidance in Text-to-Image Synthesis}
In terms of imposing spatial constraints, there is another category of methods specifically designed for controlling the contours, shapes, or layouts of objects.
ControlNet \cite{zhang2023adding} trains a new branch that incorporates spatial constraints as input and injects them into each layer of the diffusion module for customized synthesis. 
Prompt-to-prompt \cite{hertz2022prompt} enables object-specific editing by replacing the attention map in the cross-attention module. 
GLIGEN \cite{li2023gligen} designs a gated self-attention layer to incorporate spatial conditions, such as bounding boxes.
Layout-control \cite{chen2023training} employs a training-free approach that ensures higher activation values of the attention maps within the bounding box regions.
ReCo \cite{yang2023reco} achieves layout control by encoding regional tokens as part of the text prompt.
The spatial inversion module in our proposed method draws inspiration from these methods in terms of controlling the layout.
However, these methods are not developed for inversion purpose but rather assume the constrains as a prior, while our focus is on automatically discovering the underlying spatial distribution without any user specifications.

\section{Method}

\subsection{Preliminaries}
By taking an encoder-decoder view that is similar to that of variational autoencoders (VAEs) \cite{KingmaW13}, it is straightforward to inspect diffusion models.
%
%
The encoding is more commonly called a forward process that iteratively ``diffuses'' a sequence of Gaussian noises $(\mathbf{\epsilon}_t)_{t=1}^T\sim \mathcal{N}(\mathbf{0},\mathbf{I})$ into an image $\mathbf{x}_0$ using a Markov chain of $T$ steps, producing a sequence of noisy samples $(\mathbf{x}_t)_{t=1}^T$ with
\begin{equation}
    \mathbf{x}_t=\sqrt{\alpha_t} \mathbf{x}_{t-1}+\sqrt{1-\alpha_t}\mathbf{\epsilon}_t, \hspace{0.05in}1\leq t\leq T,
    \label{eq:diff_iter}
\end{equation}
where ${\alpha_t}$ controls the variance of the Gaussian noises $\mathbf{\epsilon}_t$.
%
%
%
It also defines a Gaussian distribution $q(\mathbf{x}_t\vert \mathbf{x}_{0})$ that we can use to sample latent representations for $\mathbf{x}_t$ in the generation.

The decoding is more commonly referred to as a reverse diffusion process, in which the goal is to learn another Gaussian distribution $q(\mathbf{x}_0\vert \mathbf{x}_{t})$ so that we can reconstruct $\mathbf{x}_{0}$ from $\mathbf{x}_{t}$.
Since the Markov encoding is non-reversible, the reverse diffusion is implemented by approximating $q(\mathbf{x}_0\vert \mathbf{x}_{t})$ using a model $f$ (e.g., a neural network) which is parameterized on $\mathbf{\theta}$ and learns an estimated distribution $p_\mathbf{\theta}$. 
This can be formulated as a $T$-step ``denoising'' process where, at the $t^{th}$ step, it tries to reconstruct $\mathbf{x}_{0}$ by removing noise from $\mathbf{x}_{t}$ and results in an estimation 
\begin{equation}
    \hat{\mathbf{x}}_0=f(\mathbf{x}_{t};\mathbf{\theta})\sim p_\mathbf{\theta}(\mathbf{x}_{0}\vert\mathbf{x}_{t}).
    \label{eq:diff_reconst}
\end{equation}
The learning of the model can be done based on the loss of the estimation $\hat{\mathbf{x}}_0$ from ${\mathbf{x}}_0$.
%
%
To implement text-to-image synthesis, text embedding $\mathbf{e}$ will also be fused with $\mathbf{x}_{t}$ to generate a conditioned image using Eq.~(\ref{eq:diff_reconst}) as $\hat{\mathbf{x}_0}=f(\mathbf{x}_{t}\circ\mathbf{e};\mathbf{\theta})$ where $\circ$ is a reserved fusion operator which is implemented differently in various models.
The loss is then written

\begin{align}
    L_{rec} &= \mathbb{E}\left[w_{t}\| \hat{\mathbf{x}_0}- \mathbf{x}_0\|_{2}^2\right], \nonumber \\
      &= \mathbb{E}\left[w_{t}\| f(\mathbf{x}_{t}\circ\mathbf{e};\mathbf{\theta})- \mathbf{x}_0\|_{2}^2\right],\\
      &= \mathbb{E}\left[w_{t}\| f((\sqrt{\bar{\alpha}_t} \mathbf{x}_{0}+\sqrt{1-\bar{\alpha}_t}\mathbf{\epsilon}_0)\circ\mathbf{e};\mathbf{\theta})- \mathbf{x}_0\|_{2}^2\right], \nonumber
      \label{eq:recons_loss}
\end{align}
where $w_t$ is a time-step dependent weight.

%
%
As the generated content is indeed controlled by the only input $\mathbf{e}$,
%
we can generate desired content as long as we know its text embeddings.
This is easy for pretrained concepts (e.g., \textit{cat}) because the learners have seen enough samples during the training, but hard for specific concepts (e.g., my own cat).
To address this issue, Textual Inversion is a method to backtrack the text embeddings of specific concepts \cite{gal2022image}.
%
It feeds a few samples of the target concept (e.g., 3–5 images of the user’s cat) and updates the pseudo-concept embedding ($\mathbf{e}_{\textit{cat*}}$).
%
%
It is formulated as 
\begin{equation}
    \mathbf{e}_*=\argmin_{\mathbf{e}} L_{rec}.
    \label{eq:txt_inv}
\end{equation}


\begin{table*}[t]
    \centering 
    \setlength\extrarowheight{1pt}
    \begin{tabular}{l|ccc|cc}
    \hline
        ~ & \multicolumn{3}{c|}{Comp. w/ Pretrain Concepts} & \multicolumn{2}{c}{Comp. w/ Inverted Concepts}   \\ \cline{2-6}
        Methods & Text-Align. & CoI Likelihood & Image-Align. & Text-Align. & Image-Align.  \\ \hline
        Textual Inversion \cite{gal2022image} & 0.603  & 0.032  & \textbf{0.784}  & 0.606  & 0.656   \\ 
        \quad+ Semantic Inversion & 0.645  & 0.121  & 0.762  & 0.633  & \textbf{0.664}   \\ 
        \quad+ Spatial Inversion & 0.631  & 0.116  & 0.749  & 0.620  & 0.645   \\ 
        \quad+ Semantic + Spatial & \textbf{0.702}  & \textbf{0.284}  & 0.732  & \textbf{0.662}  & 0.658   \\ \hline
        Custom Diffusion \cite{kumari2023multi} & 0.695  & 0.226  & \textbf{0.802}  & 0.702  & \textbf{0.700}   \\ 
        \quad+ Semantic Inversion & 0.701  & 0.352  & 0.760  & \textbf{0.706}  & 0.681   \\ 
         \quad+ Spatial Inversion & \textbf{0.738}  & 0.425  & 0.727  & 0.689  & 0.652   \\ 
         \quad+ Semantic + Spatial &0.734  & \textbf{0.459}  & 0.683  & 0.703  & 0.628   \\ \hline
        DreamBooth \cite{ruiz2023dreambooth} & 0.716  & 0.431  & \textbf{0.734}  & 0.691  & \textbf{0.695}   \\ 
         \quad+ Semantic Inversion & 0.720  & 0.436  & 0.718  & 0.704  & 0.683   \\ 
         \quad+ Spatial Inversion & 0.750  & \textbf{0.534}  & 0.657  & 0.705  & 0.632   \\ 
         \quad+ Semantic + Spatial & \textbf{0.753}  & 0.529  & 0.646  & \textbf{0.710}  & 0.616   \\ \hline
    \end{tabular}
    \caption{Evaluation of performance by composing with pretrained and inverted concepts, with ablation of semantic and spatial inversion components. The best results are in bold font.} 
    \label{tb:performance}
\end{table*}

\subsection{Semantic Inversion}
%
As visualized in Fig.~\ref{fig:TSNE}, Textual Inversion will make the new (pseudo-)embeddings OOD and incompatible to other concepts in the embedding space, because it does not have enough interactions with others during the post-training learning.
Our idea is then straightforward to improve its interactions for better compositionality.

To this end, we select a set of general concepts as anchors (e.g., \textit{dog}, \textit{car}, \textit{chair}, \textit{building}) and collect their text embeddings $\{\mathbf{e}_{anc}\}$. 
These concepts can be found from existing benchmark dataset like COCO \cite{lin2014microsoft} and even be combined for a wider coverage of semantic references \cite{deng2009imagenet,wei2012coaching,wei2011coached}.
We will use the anchor concepts as attractors to guide the search of the pseudo-embedding towards the core distribution.
A loss regularization is introduced as
\begin{equation}
    L_{anc}=\mathbb{E}\left[\frac{1}{\|\{\mathbf{e}_{anc}\}\|}\sum_{\mathbf{e}_i\in\{\mathbf{e}_{anc}\}}\hspace{-0.12in}\delta_i\|\mathbf{e}-\mathbf{e}_i\|_2^2\right],\\
    \text{ r.s.t }\|\mathbf{\delta}\|_0<c
\end{equation}
where $\mathbf{\delta}=\{\delta_i\}$ is a weighting vector to control the strength of the $i^{th}$ attractor, and the constraint $\|\mathbf{\delta}\|_0<c$ limits the number of active attractors to $c$.
This is to avoid the distraction from irrelevant attractors.
For example, when searching for a pseudo-embedding for a cat related concept, active attractors like \textit{cat}, \textit{pet} are preferred over irrelevant ones like \textit{car}, \textit{airplane}.
We implement the weighting vector $\mathbf{\delta}$ using sparse coding \cite{olshausen1996emergence}.

Once a pseudo-word \textit{S*} has been assigned with the embedding $\mathbf{e}_*$, it can be used in the same way as a real word in a prompt for image generation (e.g., ``\textit{a S* cat sitting next to a dog}'').
Each word in a prompt is more commonly referred to as a token.
The model first assigns an embedding for each token and feeds these embeddings into a text Transformer where they are refined into the actual token embeddings that will be used as conditions in the generation (or reverse) process.
To simplify the description, we will still use the symbol $\mathbf{e}$'s to represent these refined token embeddings.

\subsection{Spatial Inversion}
To make the image generation conditioned on the token embeddings in the reverse process, a popular way is to use transformer blocks.
More specifically, an attention map will be calculated for each token embedding to indicate its appearance or how it is attended in the resulting image (e.g., location, shape, details).
This is implemented by the cross-attention mechanism as
\begin{equation}
    \mathbf{A}_i=softmax\left(\frac{\phi(\mathbf{x}_t)\kappa(\mathbf{e}_i)^\top}{\sqrt{d_k}}\right),
\end{equation}
where $\phi$ and $\kappa$ are the image feature extractor and text feature extractor respectively and $d_k$ is the dimensionality of $\kappa(\mathbf{e}_i)$.
Several pioneer works have found that the appearance of the token can be controlled by manipulating this attention map \cite{hertz2022prompt,parmar2023zero,chen2023training}.
Therefore, we can regulate the attention maps to be attended on the right tokens to avoid the situation that the pseudo tokens dominate the generation process.
In spatial inversion, we propose a method to recover the coherent locations of a pseudo token (e.g., \textit{S*}) and concepts being combined (e.g., \textit{dog}) in a prompt.
The locations are then used to regulate the attention maps of tokens.

\begin{figure*}[ht]
\centering
\includegraphics[width=0.85\textwidth]{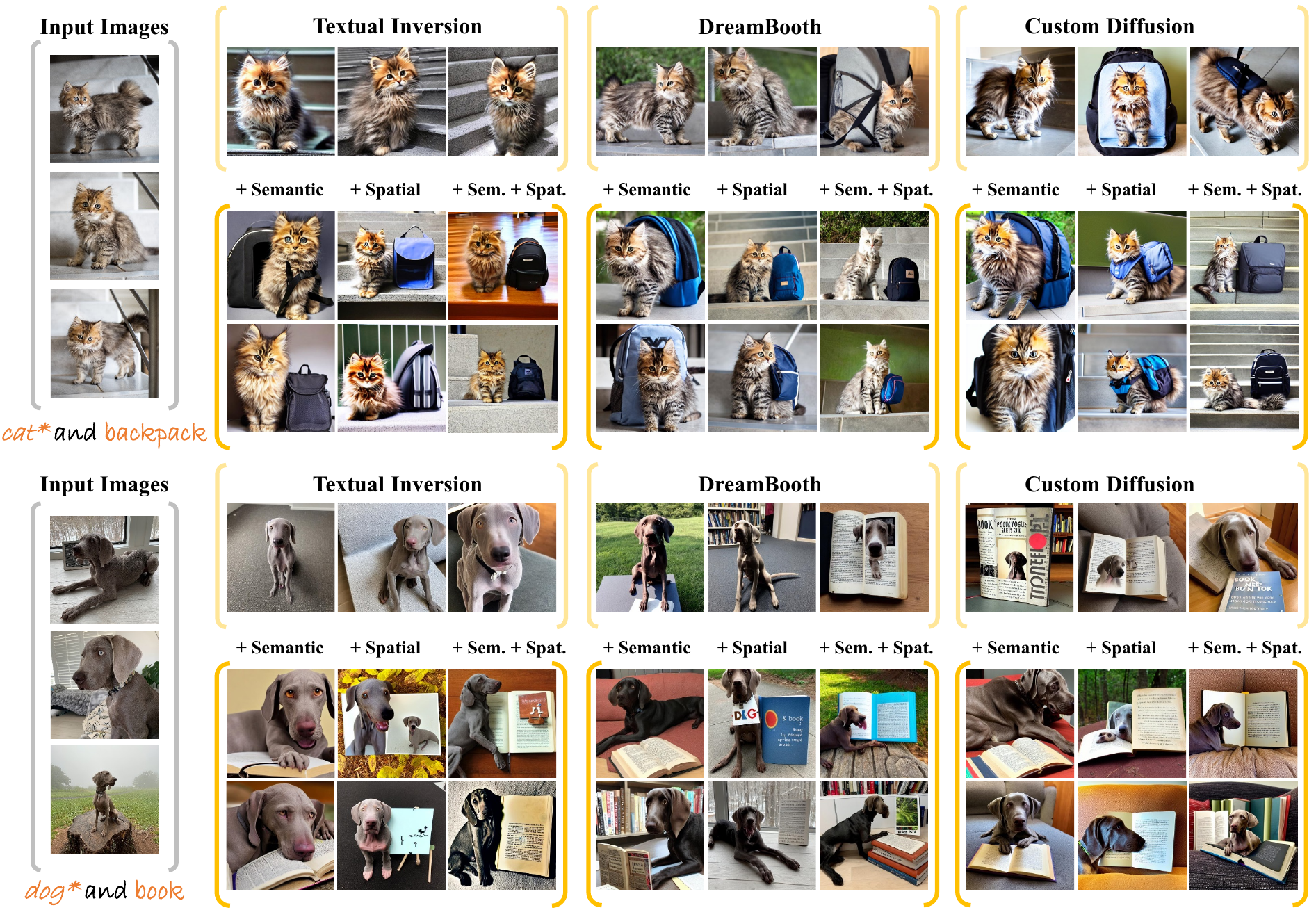} 
\caption{Examples of composing inverted concepts \textit{cat*} and \textit{dog*} with pretrained concepts \textit{backpack} and \textit{book}.}
\label{fig:with_pretrain}
\end{figure*}

We implement the location recovery by training an $MLP$ model which takes two token embeddings as the input and outputs the locations as
\begin{equation}
    \mathbf{l}_i,\mathbf{l}_j=MLP(\mathbf{e}_i,\mathbf{e}_j),
\end{equation}
where $\mathbf{l}_i,\mathbf{l}_j\in \mathbb{R}^4$ are coordinates of the bounding boxes of the attended areas of the two tokens.
We simplify the method by only considering noun tokens and construct a vocabulary of frequently used nouns \cite{lin2014microsoft}.
The nouns are then combined as prompts and used to generate images.
The object detection \cite{carion2020end} is conducted on the resulting images to find the bounding boxes of these nouns which serve as the ground truth for the training.
For tokens in the vocabulary, we assign it to the noun that is the nearest one in the embedding space.
The $MLP$ is then able to recover the locations of any given tokens.

With the location bounding boxes, we convert each of them into an attention mask $\mathbf{M}_i$ which is with the same size as that of $\mathbf{A}_i$.
We can then manipulate the attention maps by introducing a location regularization loss as
\begin{equation}
    L_{loc} = \frac{1}{N}\sum_{i=1}^{N}\left(1 - \frac{\sum \left(\mathbf{M}_{i} \circ \mathbf{A}_{i}\right)}{\sum \mathbf{A}_{i}}\right).
    \label{eq:loss_loc}
\end{equation}
It encourages the tokens to be attended on the locations indicated by the masks and penalizes deviations. 
It is calculated at the first 10 reverse steps to update the latent variable $\mathbf{x}_t$.

\begin{figure*}[ht]
\centering
\includegraphics[width=0.84\textwidth]{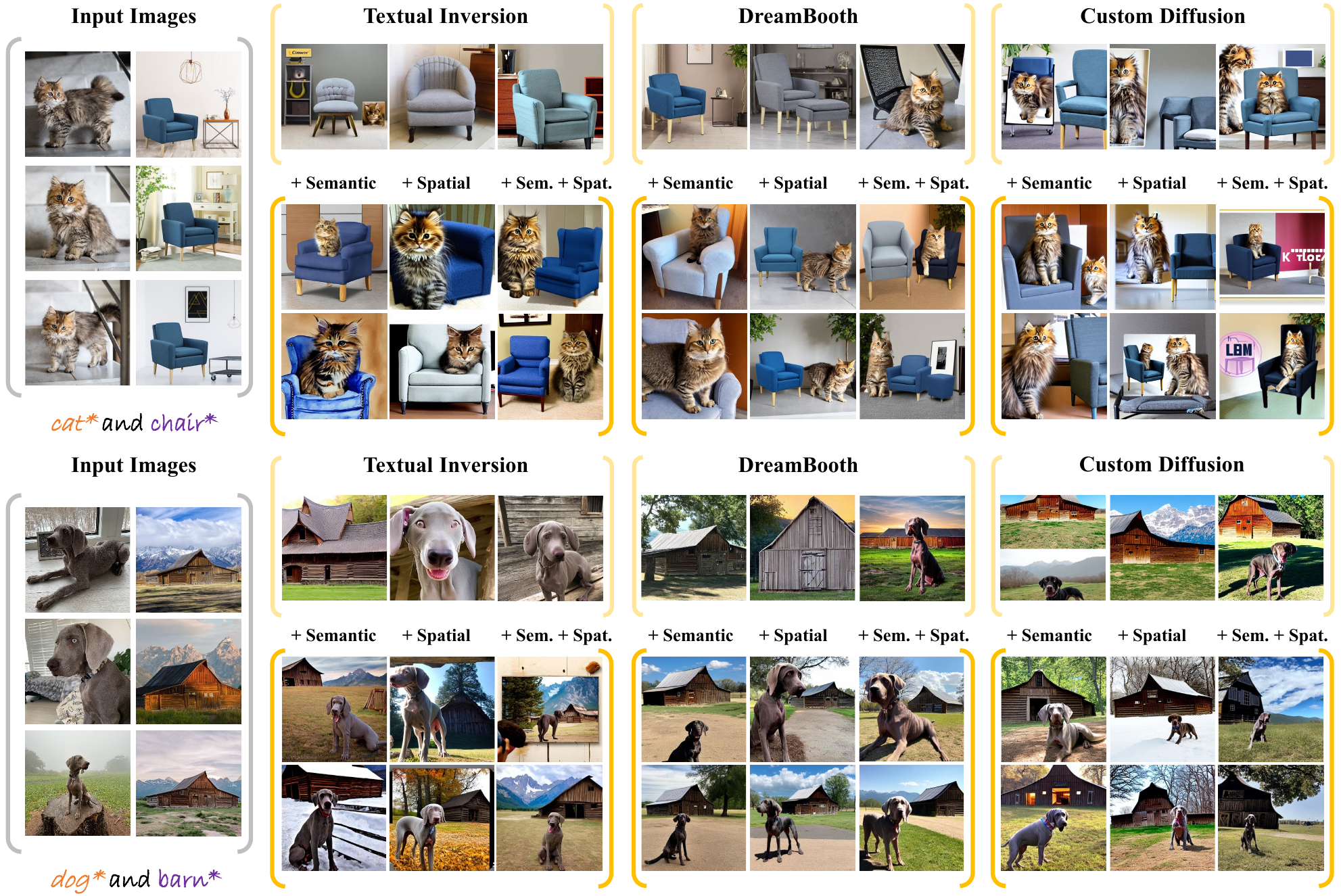} 
\caption{Examples of composing inverted concepts of \textit{cat*}, \textit{chair*}, \textit{dog*}, and \textit{barn*} to each other.}
\label{fig:with_invt}
\end{figure*}
\begin{figure*}
\centering
\includegraphics[width=0.96\textwidth]{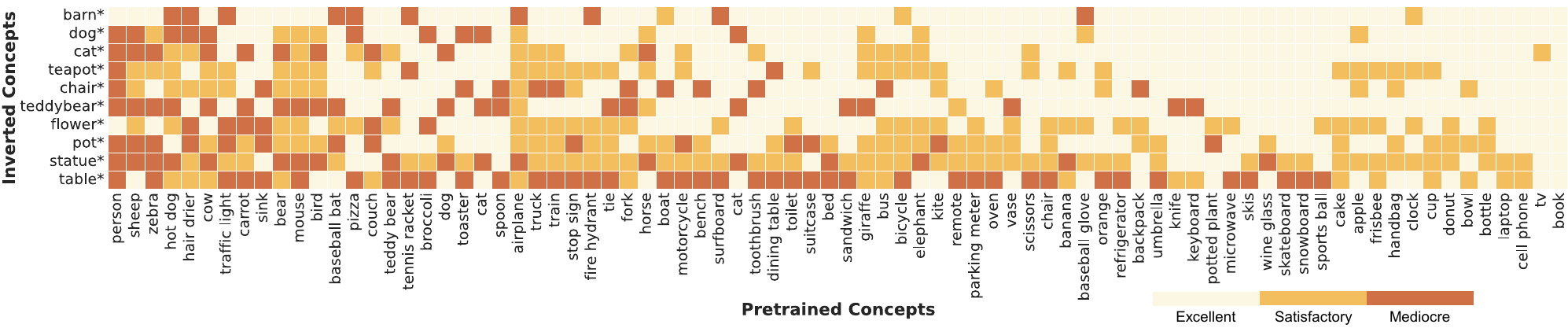} 
\caption{Assessment of compositional synthesis quality through user evaluations.}
\label{fig:compositionality}
\end{figure*}

\section{Experiments}
To evaluate the performance of our proposed methods, we conduct experiments by combining the inverted concepts with both pretrained and inverted concepts.

\textbf{Datasets.}
We construct a comprehensive dataset by accumulating almost all open-sourced concepts used in previous studies \cite{kumari2023multi,gal2022image,ruiz2023dreambooth}. 
It consists of 10 concepts of 2 animal, 2 furniture, 2 object/container, 1 house, 1 plant, 2 toy categories. 
To test the generalizability, we generate prompts by combining the inverted concepts with 80 categories from the COCO dataset \cite{lin2014microsoft} using the conjunction word ``and''. 
This results in 1600 prompts and generates 16000 images.
%
%
%
We also combine the inverted concepts to each other, resulting in 90 prompts and 900 images generated.
This is also aligned to the multi-concept composition task in previous studies.

\textbf{Evaluation Metrics.}
We utilized three evaluation metrics: 1) \textit{Text-alignment} which quantifies the extent to which a generated image accurately represents the semantics of the text prompt, as determined by the CLIP similarity \cite{radford2021learning}.
%
%
2) \textit{CoI Likelihood} which measures the probability that CoIs present in the results using an object detector (DETR \cite{carion2020end} based on ResNet101 \cite{he2016deep} and pretrained on the COCO dataset). 
%
%
3) \textit{Image-alignment} which evaluates the extent to which the generated images are visually similar to the user samples, as determined by the cosine similarity of their CLIP image features.

\textbf{Baselines.} We employ 3 poplar state-of-the-art (SOTA) methods as the baselines including
1) \textit{Textual Inversion (TI)} \cite{gal2022image} which focuses on fine-tuning the text embedding exclusively. We employ the Stable Diffusion version, using the parameters reported by the authors in their paper. 
2) \textit{DreamBooth} \cite{ruiz2023dreambooth} which fine-tunes all parameters of the U-Net architecture. As DreamBooth does not fine-tune the text embedding, we integrate TI into DreamBooth to apply the proposed method in this paper. 
3) \textit{Custom Diffusion} \cite{kumari2023multi} which aims to fine-tune partial parameters in the cross-attention modules. For the composition of inverted concepts, we adopt the joint training strategy, as it has been highlighted in the paper as the best-performing approach. 
The third-party implementations from HuggingFace are used for all the aforementioned methods. 
In the fine-tuning and inference stages, we follow the usual practice to use the \textit{S*} and superclass token to represent the inverted concept (e.g., ``\textit{cat* cat}''). 
%

\subsection{Performance} 
The results are presented in Table~\ref{tb:performance}.
%
The proposed method exhibits improvements over SOTA methods in terms of 16.4\% (787.5\%), 5.6\% (103.1\%), 5.2\% (22.7\%) on Text-Align (CoI Likelihood) compared to TI, Custom Diffusion, and DreamBooth, respectively. 
There is only a slight trade-off of 6.6\%, 14.8\%, and 12.0\% on Image-Align when compared to the three methods.
The performance gain on CoI Likelihood reaches 52.9\% when composing with pretrained concepts, indicating a significant improvement.
%
%
Another observation is that the augmented TI achieves a comparable performance to the original Custom Diffusion and DreamBooth. 
This is surprising because SOTA performance is achieved without any fine-tuning of network parameters.

Fig.~\ref{fig:with_pretrain} shows two examples of composing inverted concepts with pretrained concepts.
The proposed method clearly improves the performance in the presence of the pretrained concepts.
Note that the semantic inversion module primarily emphasizes semantic completeness, occasionally resulting in the generation of low-probability scenes (such as half a cat in a backpack or a dog reading a book). 
On the other hand, the spatial inversion module tends to generate scenes that align with more common statistical occurrences.

Fig.~\ref{fig:with_invt} presents two examples of composing the inverted concepts to each other.
The presence of the CoIs is also significantly increased.
A noticeable difference compared to the results in Fig.~\ref{fig:with_pretrain} is the larger variation in the appearance of the concepts of interest. Specifically, the generated cats, dogs, and barns exhibit a wider range of viewpoints.

\subsection{User Study}
To assess the computational efficiency and quality of the synthesis, we conducted a user study. 
We randomly selected 1,600 images generated by the proposed method and enlisted the participation of two users to rate the synthesis quality. 
The ratings were divided into three categories: \textit{Excellent} represents the successful generation of two CoIs without any unnatural details; \textit{Satisfactory} indicates that the CoIs were generated in an acceptable manner, though some minor flaws may be present; and \textit{Mediocre} signifies the presence of obvious unreasonable details or missing CoIs.
The results of the user study are presented in Fig.~\ref{fig:compositionality}.
The high capability of the proposed method in generating quality images is clearly evident, as indicated by a probability of 81.9\% for receiving ratings above the \textit{Satisfactory}.

Additionally, we make the assumption that the quality rating serves as an indicator of compositionality. In other words, when the probability of generating high-quality images through the composition of two concepts is higher, it suggests that those concepts are easier to compose.
In Fig.~\ref{fig:compositionality}, it becomes apparent that rigid objects (such as $book$ and $tv$) are more straightforward to compose. This observation is supported by the fact that 9 out of the 10 rightmost concepts in the figure are rigid objects. 
This finding aligns with our understanding as rigid objects possess more consistent appearances and visual characteristics.
In contrast, non-rigid objects like animals (e.g., $cow$, $sheep$) are challenging to compose, as indicated by the fact that 7 out of the 10 leftmost concepts are non-rigid objects.


\section{Conclusion}
We have identified the mechanism that causes the overfitting and dominance of the inverted concepts in generation.
To address the issue, we propose a compositional inversion method which consists of two modules of semantic and spatial inversions.
The semantic inversion guides the inversion towards the core distribution to ensure better coherence with other concepts, while the spatial inversion discovers the underlying layout distribution for CoIs and uses it to regularize the attention maps.
The experimental results have validated the effectiveness of the method.

\section{Acknowledgments}
Joint support for this research was provided by the Hong Kong Research Grants Council through the General Research Fund (Project No.: 15200023), the National Natural Science Foundation of China (Grant No.: 62372314), and InnoHK program.

\bibliography{aaai24}

\clearpage
\appendix
\section{Appendix}
To provide a more comprehensive understanding of the method, we have included additional details in the following sections.
\textbf{The source code can be accessed at {\color{blue} https://github.com/zhangxulu1996/Compositional-Inversion}.}

\subsection{Additional Implementation Details}
%
We utilize the third-party implementation of HuggingFace for all state-of-the-art (SOTA) methods. 
The Stable Diffusion model serves as our pretrained model. 
To ensure a fair comparison, we employ 50 steps of DDPM sampling with a scale of 7.5 for all methods. 
All experiments are conducted using A-100 GPUs. 
During the fine-tuning and inference stages, we adhere to the conventional practice of representing the inverted concept using the \textit{S*} and superclass token (e.g., ``$cat*$ $cat$'').

\textbf{Textual Inversion} 
We train with the recommended batch size of 8 using a learning rate of 0.005 (scaled by batch size for an effective learning rate of 0.04) for 5,000 steps.

\textbf{Custom Diffusion} 
We train with a batch size of 2 using a learning rate $1 \times 10^{-5}$ (scaled by batch size for an effective learning rate of $2 \times 10^{-5}$) for 500 steps. 
We collect 200 images with the same class as the inverted concept as the regularization dataset to avoid language drift.

\textbf{DreamBooth} As DreamBooth does not fine-tune the text embedding, we integrate Textual Inversion into DreamBooth to apply the proposed method in this paper. 
The parameters of the text Transformer are frozen and the U-Net diffusion model is fine-tuned with a learning rate $1 \times 10^{-5}$ and the batch size of 1 for 800 steps.

\textbf{Compositional Inversion (Ours)}
To balance $L_{rec}$ and $L_{anc}$, the weights are set to 0.05 and 0.95 respectively in Textual Inversion and Custom Diffusion. In Dreambooth, the weight of $L_{rec}$ is 0.4 and $L_{anc}$ is 0.6. The balance of the losses are controlled by a factor $\lambda$.

\subsubsection{Balance of the Reconstruction Loss $L_{rec}$ and Semantic Regularization $L_{anc}$}
In accordance with the specifications outlined in the \textbf{Method} section, we incorporate two distinct types of losses during the inversion process: the reconstruction loss ($L_{rec}$) and the semantic regularization loss ($L_{anc}$). To achieve a balance between these two losses, we employ a weight parameter denoted as $\lambda$, which results in
\begin{equation}
L_{final} = (1-\lambda)L_{rec} + \lambda L_{anc}
\end{equation}

\begin{figure}[t]
\centering
\includegraphics[width=0.9\columnwidth]{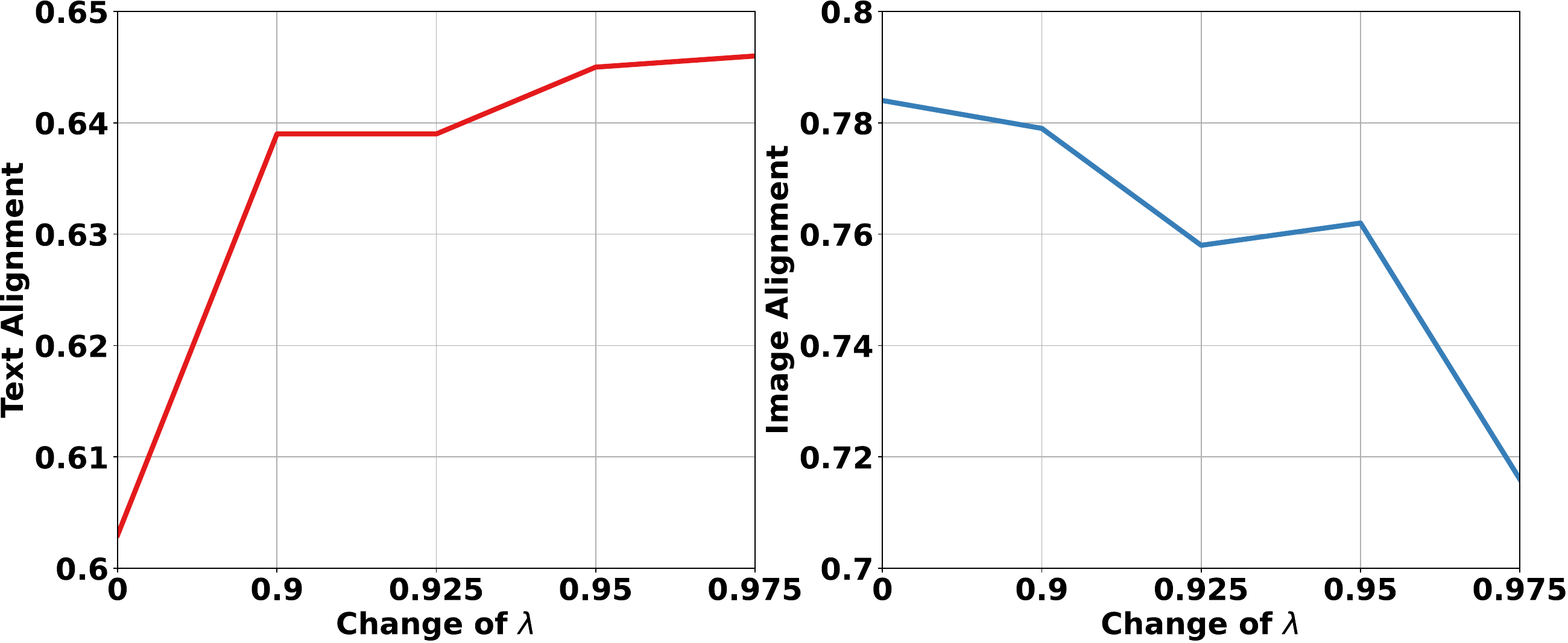} 
\caption{The development of the semantic fidelity (Text Alignment) and image fidelity over the variation of the $\lambda$.}
\label{fig:lambda}
\end{figure}

\begin{figure}[t]
\centering
\includegraphics[width=0.9\columnwidth]{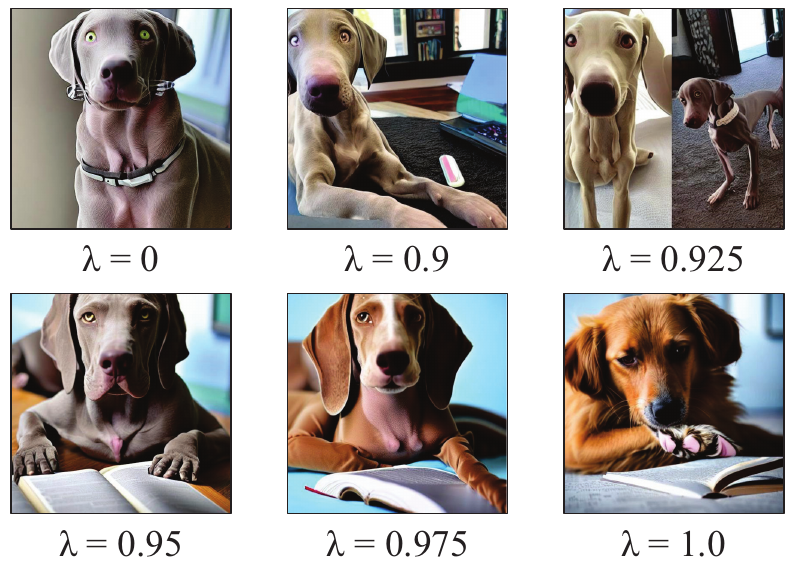} 
\caption{Images generated with different $\lambda$ values for the given prompt ``\textit{dog* and book}''.}
\label{fig:lambda_samples}
\end{figure}

To determine the optimal configuration, a series of experiments are conducted. Specifically, we fine-tune the $\lambda$ parameter within the range of $[0,1]$ using a step size of $0.25$. 
During this process, we observe the development of both the semantic fidelity to the prompts (Text Alignment) and the fidelity to the user images (Image Alignments). 
The results are visualized in Figure~\ref{fig:lambda}. 
The optimal balance between the two alignments is found to occur at $\lambda = 0.95$. 
This can be observed in the provided example depicted in Figure~\ref{fig:lambda_samples}, where both the semantic and image fidelity are effectively balanced at $\lambda = 0.95$.

\subsubsection{Spatial Inversion for Location Regularization}
\begin{figure*}[t]
\centering
\includegraphics[width=0.96\textwidth]{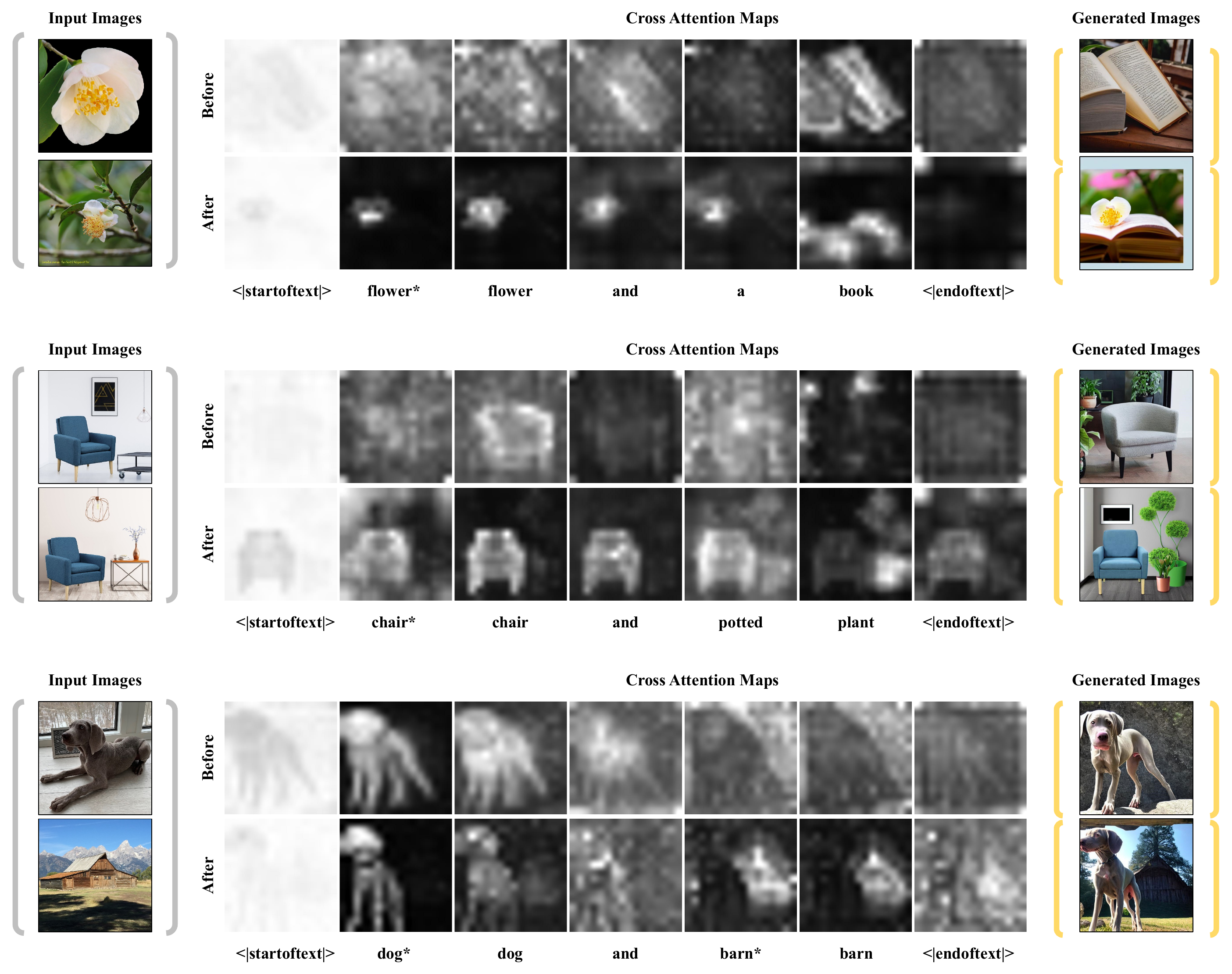} 
\caption{Visualization of attention maps before and after the application of Spatial Inversion.}
\label{fig:layout}
\end{figure*}

For a more intuitive illustration of the effect of imposing Spatial Inversion.
In Fig.~\ref{fig:layout}, We have visualized how the attention maps have affected. 
It is evident that the dominance of the $dog*$ over $barn*$ has been eliminated by using the Spatial Inversion as regularization.

\subsection{Further on the Concept Compostionality}
We assess the quality of generation to evaluate the concept compositionality.
However, the compositionality is defined over the inverted concepts and the image alignment to the user samples.
This is indeed an instance-level compositionality.
In this section, we are interested in the concept compositionality in a more general sense.
To this end, we  relax the image alignment and evaluate how the semantics of the prompts have been implemented in the resulting images.
In other words, we evaluate the compositionality at a class level.
The specific questions we are trying to answer include
\begin{itemize}
    \item Are the pretrained concepts compositional to each other or how much they are compositional when combined in the same prompt?
    \item How much the inverted concepts are compositional to the pretrained concepts?
\end{itemize}

The experiment is conducted on 4005 concept pairs of a set of 90 concepts including 80 pretrained and 10 inverted concepts.
For each pair, we generate 10 prompts including the two concepts, resulting in 40050 prompts.
The text alignment is evaluated for each prompt on the generated image.
The mean of the 10 text alignment is used as the compositionality of the pair.

\begin{figure*}[t]
\centering
\includegraphics[width=0.9\textwidth]{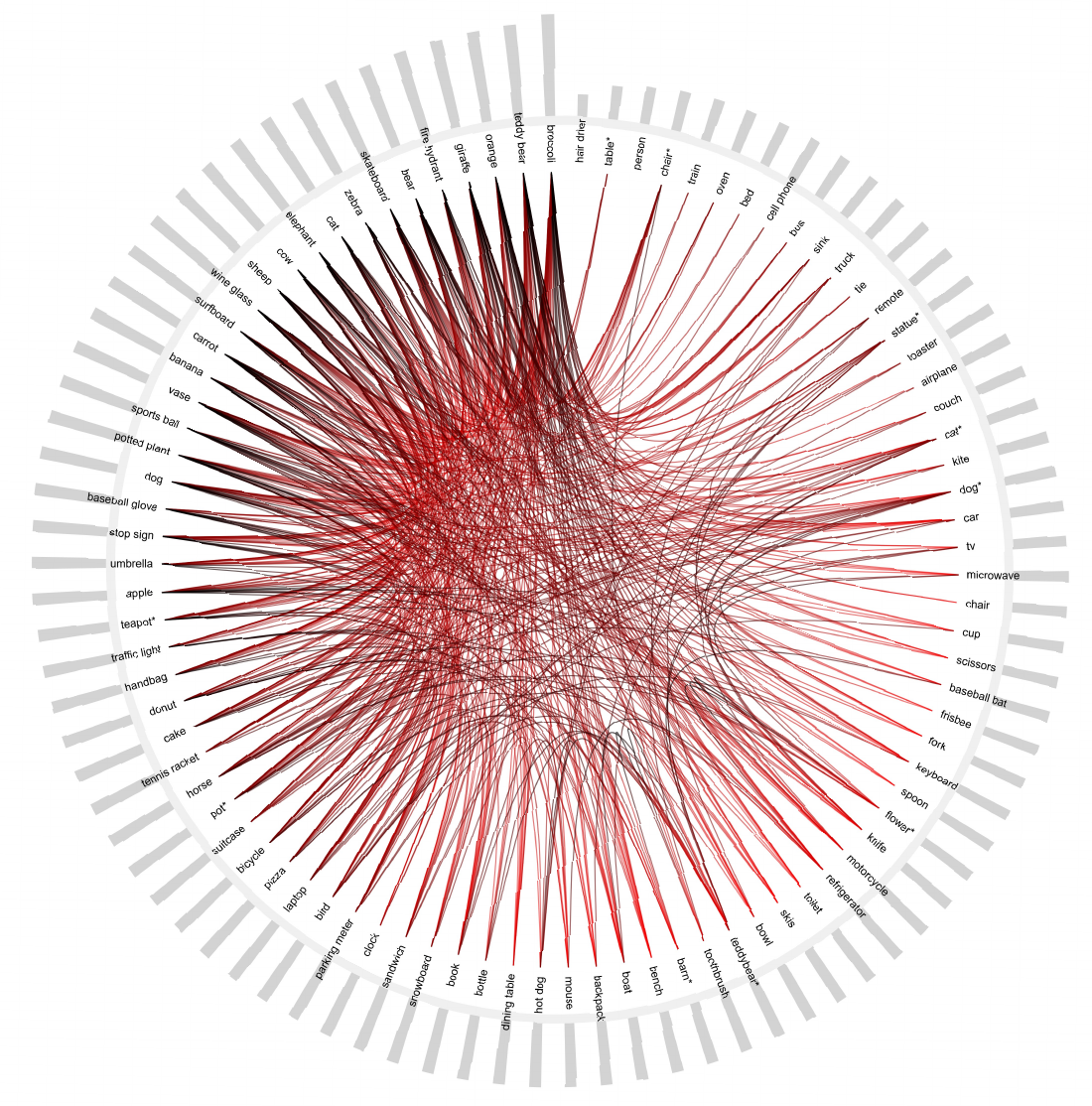} 
\caption{A visualization of the concept compositionality. The mean compositionalities of concepts are shown as outer bars, while the inter-concept compositionalities are shown as inner links. A threshold of 0.81 is used to filter out the links.}
\label{fig:circos}
\end{figure*}

A visualization of the results is shown as Fig.~\ref{fig:circos}.
It is evident that the compositionality vary from concepts.
Hypothetically, the top-ranked concepts tend to possess unique features, such as the distinct color and texture of $broccoli$ or the unique color and shape of an $orange$.
Conversely, concepts with lower rankings may exhibit greater variation in their appearances.
Interestingly, the inverted concepts appear to be intertwined with the pretrained concepts, suggesting a convergence of their compositionality towards a similar level as the pretrained ones.

\subsection{Additional Results}
\begin{figure*}[ht]
\centering
\includegraphics[width=0.98\textwidth]{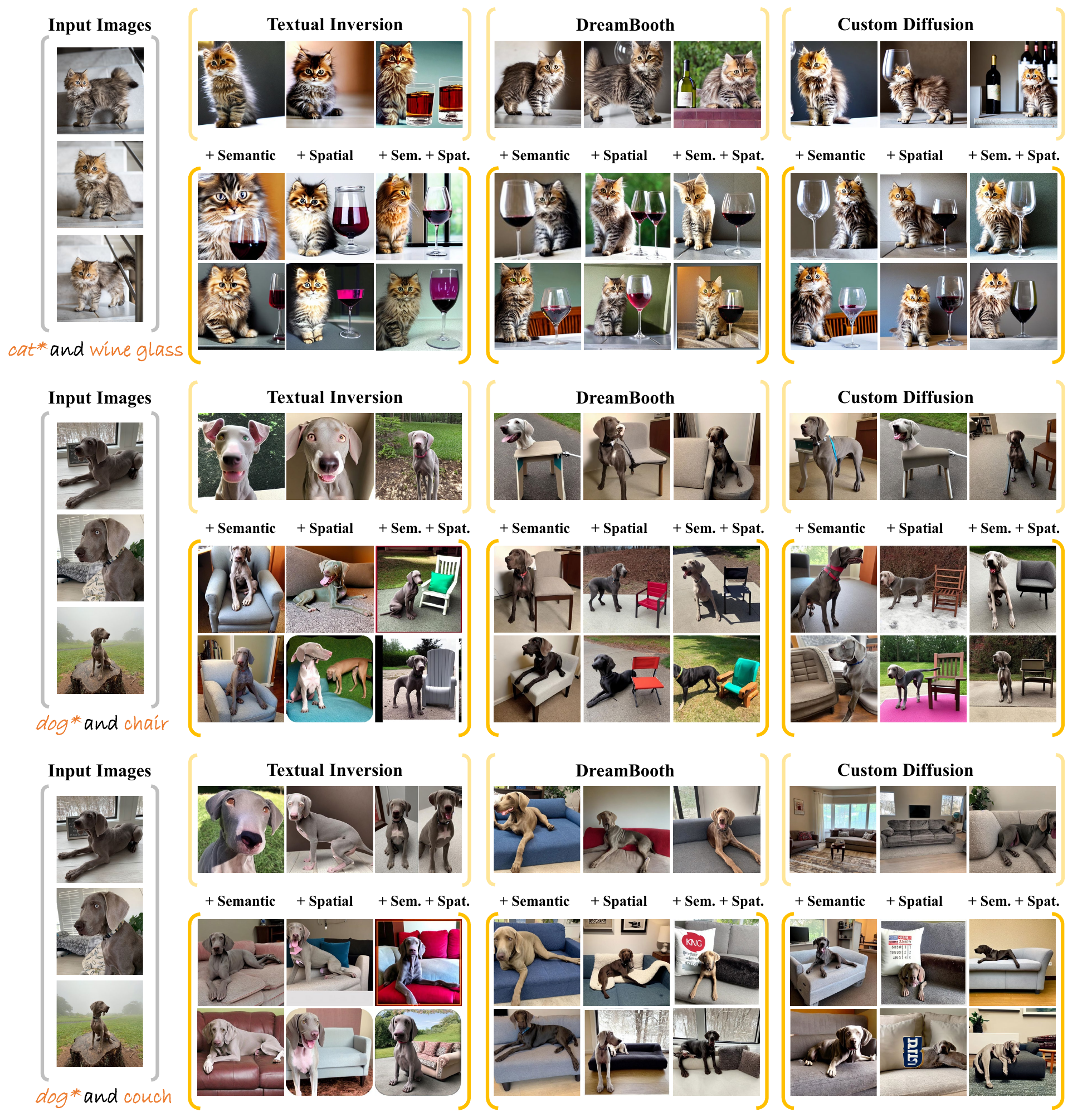} 
\caption{Examples of composing inverted concepts with pretrained concepts. Images in the light-orange brackets are generated with traditional inversion methods, while the ones in the orange brackets are with our proposed Compositional Inversion.}
\label{fig:with_pretrain_supple}
\end{figure*}

\begin{figure*}[ht]
\centering
\includegraphics[width=0.98\textwidth]{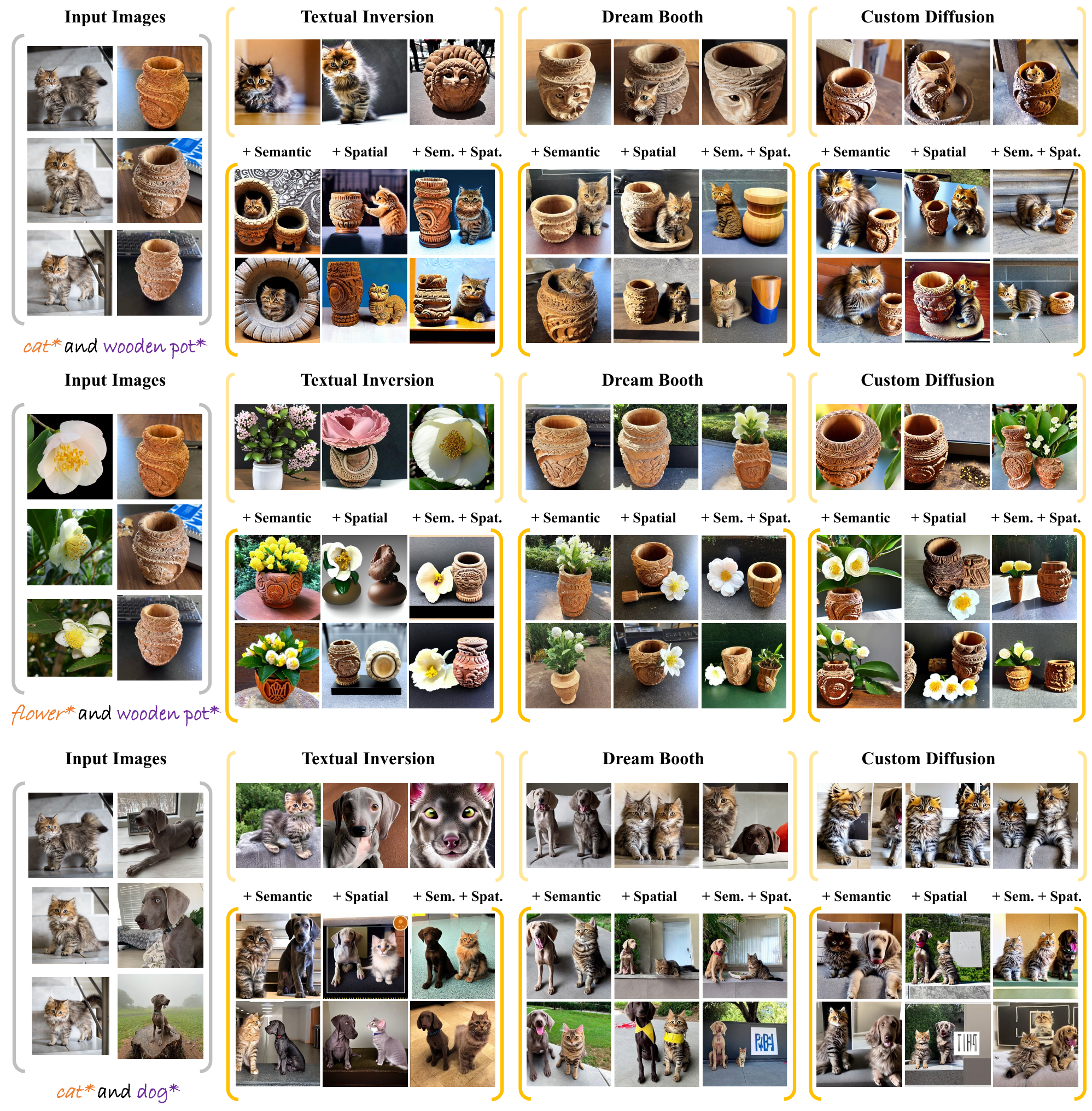} 
\caption{Examples of composing inverted concepts. Images in the light-orange brackets are generated with traditional inversion methods, while the ones in the orange brackets are with our proposed Compositional Inversion.}
\label{fig:with_inverted_supple}
\end{figure*}

\begin{figure*}[ht]
\centering
\includegraphics[width=0.98\textwidth]{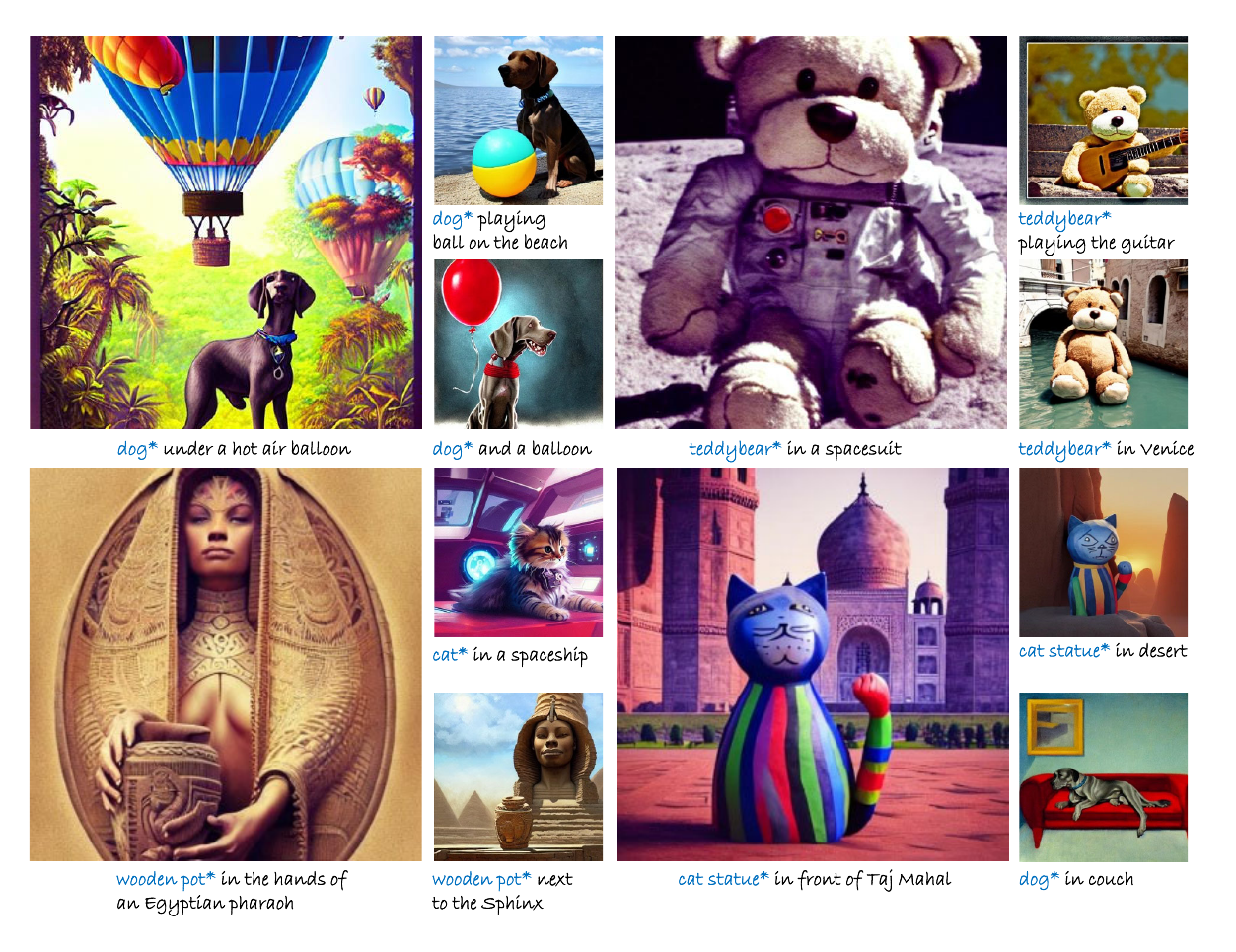} 
\caption{Examples of composing inverted concepts with concepts not in the COCO vocabulary.}
\label{fig:super_imgs_supple}
\end{figure*}

In this section, we provide more samples of the generated images.
Fig.~\ref{fig:with_pretrain_supple} shows the composition with pretrained concepts, and Fig.~\ref{fig:with_inverted_supple} shows the composition of inverted concepts. 
These figures exhibit our method's ability to generate meaningful compositions using both pretrained and inverted concepts, proving its versatility and robustness in different scenarios.
To demonstrate the generalizability, we also compose the inverted concepts with concepts not in the COCO vocabulary (e.g., $spacsuits$, $Sphinx$).
Results are presented in Fig.~\ref{fig:super_imgs_supple}.


\end{document}